\documentclass[10pt,twocolumn,letterpaper]{article}

\usepackage{iccv}
\usepackage{times}
\usepackage{epsfig}
\usepackage{graphicx}
\usepackage{amsmath}
\usepackage{amssymb}

\usepackage{booktabs}
\usepackage{multirow}
\usepackage{threeparttable}
\usepackage{pifont}
\usepackage{graphicx} 
\usepackage{subfigure} 

\newtheorem{lemma}{Lemma}

\usepackage{color}
\usepackage[misc]{ifsym}

\usepackage[breaklinks=true,bookmarks=false]{hyperref}

\iccvfinalcopy 

\def\iccvPaperID{4534} 

\ificcvfinal\fi

\begin{document}
	
	\title{Transductive Episodic-Wise Adaptive Metric for Few-Shot Learning}
	
	\author{Limeng Qiao$^{1,4}$,
		Yemin Shi$^{2,4}$, 
		Jia Li$^{3,4}$\thanks{Correspondence should be addressed to Yonghong Tian and Jia Li. (yhtian@pku.edu.cn and  jiali@buaa.edu.cn).},
		Yaowei Wang$^{4}$,
		Tiejun Huang$^{2,4}$ and
		Yonghong Tian$^{2,4*}$, \\
		$^{1}$ Center for Data Science, AAIS, Peking University \\
		$^{2}$ National Engineering Laboratory for Video Technology, School of EE\&CS, Peking University\\
		$^{3}$ State Key Laboratory of Virtual Reality Technology and Systems, SCSE, Beihang University\\
		$^{4}$ Peng Cheng Laborotory, Shenzhen, China
	}
	
	\maketitle
	\setlength{\abovedisplayskip}{3pt}
	\setlength{\belowdisplayskip}{3pt}
	
	\begin{abstract}
		Few-shot learning, which aims at extracting new concepts rapidly from extremely few examples of novel classes, has been featured into the meta-learning paradigm recently. Yet, the key challenge of how to learn a generalizable classifier with the capability of adapting to specific tasks with severely limited data still remains in this domain. To this end, we propose a \textbf{T}ransductive \textbf{E}pisodic-wise \textbf{A}daptive \textbf{M}etric (TEAM) framework for few-shot learning, by integrating the meta-learning paradigm with both deep metric learning and transductive inference. With exploring the pairwise constraints and regularization prior within each task, we explicitly formulate the adaptation procedure into a standard semi-definite programming problem. By solving the problem with its closed-form solution on the fly with the setup of transduction, our approach efficiently tailors an episodic-wise metric for each task to adapt all features from a shared task-agnostic embedding space into a more discriminative task-specific metric space. Moreover, we further leverage an attention-based bi-directional similarity strategy for extracting the more robust relationship between queries and prototypes. Extensive experiments on three benchmark datasets show that our framework is superior to other existing approaches and achieves the state-of-the-art performance in the few-shot literature.
	\end{abstract}
	
	
	\section{Introduction}
	Deep neural networks have achieved great success on many practical applications recently and even have surpassed humans in image recognition domain \cite{he2016deep, krizhevsky2012imagenet, simonyan2014very}. However, these successes largely benefit from tens of thousands of labeled data, enormous number of parameters and sophisticated training strategies. In terms of the low-data scenarios, such as medical images classification and marine biological recognition, deep neural networks will be over-fitting and severely collapse because only rare labeled samples are provided for training. To this end, by exploring the exciting idea that tries to learn new concepts rapidly and generalize well with extremely few examples, or even single, few-shot learning has attracted significant research interest recently \cite{finn2017model, munkhdalai2017meta, ravi2016optimization, santoro2016meta, snell2017prototypical, vinyals2016matching,  yang2018learning}.
	
	\begin{figure}[t]
		\begin{center}
			\includegraphics[width=0.96\linewidth]{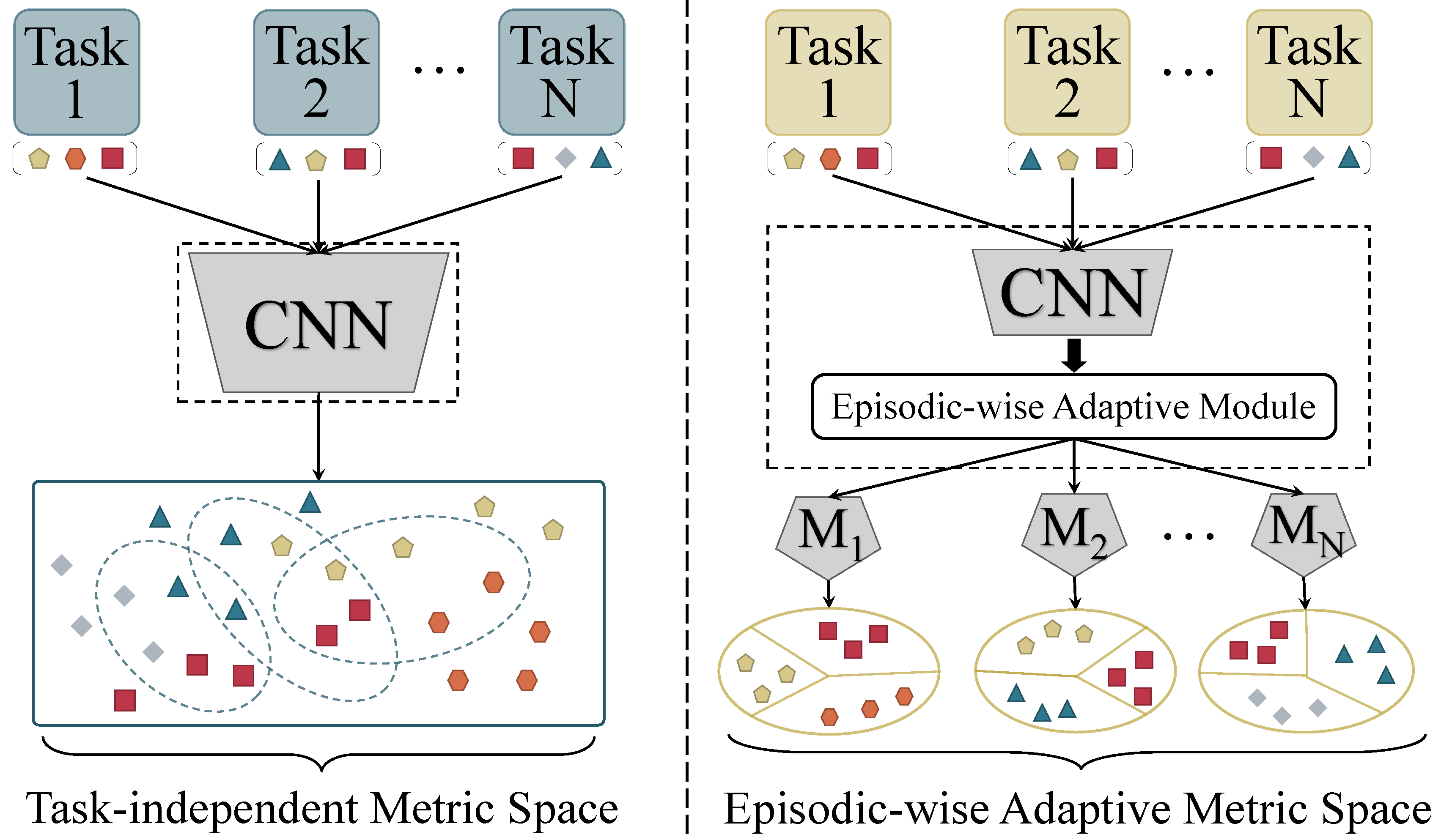}
		\end{center}
		~\\[-11mm]
		\caption{The left part shows a general framework of previous approaches based on meta learning. The model embeds all samples into a task-independent metric space with a shared CNN. However, considering the distinctive information within each task, we further learn an episodic-wise adaptive metric for classification. The right part illustrates the motivation of our approach briefly.}
		\label{fig:motivation}
		\vspace{-0.6cm}
	\end{figure}
	
	Concretely, we focus on the case of few-shot classification with the meta-learning paradigm, which leverages a series of independent and identically distributed few-shot tasks to learn a desired classifier at training time and applies the model directly to non-overlap unseen target classification problem during testing phase. The so-called episodic-training strategy, which has been widely developed in many previous work \cite{snell2017prototypical, vinyals2016matching}, keeps the consistency between the training and real test scenario and improve the generalization performances. In each task (or episode), only a handful of labeled examples per classes (the support set) are provided to the classifier for training and then plenty of unlabeled points (the query set) need to be assigned labels for prediction. Specifically, \cite{snell2017prototypical, vinyals2016matching,  yang2018learning} jointed the meta-learning with metric learning as a feed-forward manner, with optimizing a shared distance metric across all tasks. Moreover, \cite{finn2017model} proposed to learn a general initialization strategy and \cite{li2017meta, ravi2016optimization} learned to update the parameters of learner directly with a higher-level meta-optimizer (e.g. LSTM). 
	
	While these approaches based on meta-learning paradigm have made significant advances in few-shot classification, they do suffer from two distinct limitations. One is that all examples from various different tasks are embedded into a task-independent metric space indiscriminately, see Fig.\ \ref{fig:motivation} (left). Namely, this assumption does not take the task-level information (meta-data) into account but example-level feature only, which neglects the specificity of different tasks. Actually, what is missing of this idea is an adaptive module that tailors the metric space for each task. The other is that most current methods follow the setup of inductive inference, that is, training the meta-learner with severely limited support data and predicting queries one by one in each task. Obviously, this process does not adequately consider the interaction between the support set and unlabeled test set and thus weakens the advantages of meta-learning.
	
\begin{figure*}[t]
		\begin{center}
			\includegraphics[width=1.0\linewidth]{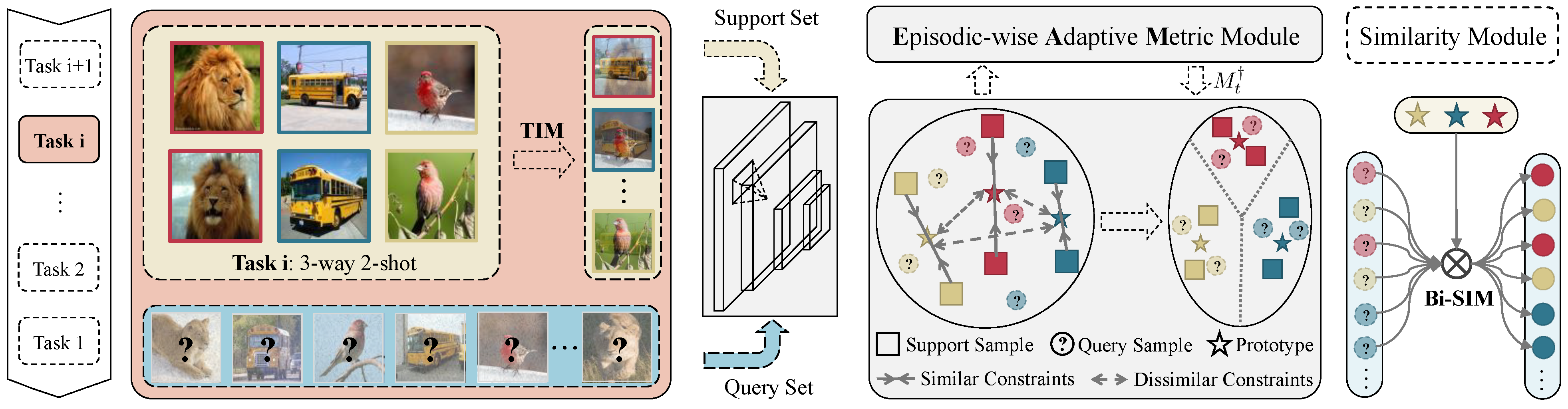}
		\end{center}
		~\\[-10mm]
		\caption{
			The architecture of \textbf{T}ransductive \textbf{E}pisodic-wise \textbf{A}daptive \textbf{M}etric (TEAM) framework for few-shot learning. The whole training data is reorganized into a pool of few-shot tasks (meta-training set) firstly and then the task-level data augmentation technique (TIM) performs the convex combination between all samples in each task. After that, the augmented episodic data is fed into the CNN and the EAM module constructs a task-specific distance metric $M_t^\dagger$ explicitly for each task on the fly with the pairwise constraints and regularization prior. The symbol $\otimes$ in Bi-SIM module indicates that each query is assigned a label using the bi-directional similarity.}
		\label{fig:framework}
		\vspace{-0.25cm}
	\end{figure*}

	To deal with the key challenge of how to learn a generalizable classifier with the capability of adapting to specific tasks with severely limited data, we propose a novel meta-learning framework for few-shot classification, named as \textit{\textbf{T}ransductive \textbf{E}pisodic-wise \textbf{A}daptive \textbf{M}etric} (TEAM), which efficiently tailors an episodic-wise metric space with applying the idea of transductive inference. Specifically, we put forward not only to learn a task-agnostic instance embedding model end-to-end over a pool of few-shot tasks as a meta-learning manner, but also constructs a task-specific distance metric explicitly for each task with the distinctive information, such as the pairwise constraints and regularization prior, see Fig. \ref{fig:motivation} (right). Furthermore, we formulate the optimization process for task-specific metric into a standard semi-definite programming (SDP) problem \cite{boyd2004convex} and finally acquire the closed-form solution by solving the SDP problem on the fly. Hereafter all features generated by the task-agnostic model are adapted into task-specific metric spaces where samples from the same class are closer and different classes are farther apart. According to the transformed embeddings, we then perform a novel attention-based bi-directional similarity strategy to compute more robust relationship between each unlabeled query and class prototypes, which further improves the performance of our approach. In addition, by utilizing the convex combination of all samples within each task to construct auxiliary training tasks, we propose a task-level data augmentation technique for boosting the generalization of the embedding model. The whole framework is illustrated in Fig. \ref{fig:framework} in details. 
	
	The main contribution is summarized as threefold. (1) We propose a general meta-learning framework of applying the transductive inference on formulating the adaptation procedure into a SDP problem and tailoring the episodic-wise metric in each task for few-shot learning, which can also be directly extended to other existing methods and even semi-supervised learning. (2) We identify a novel bi-directional similarity strategy for extracting the more robust relationship between queries and prototypes. (3) The  experimental results on three benchmark datasets show that our framework is superior to other state-of-the-art approaches. 
	
	
	
	
	
	\section{Related Work}
	
	\noindent
	\textbf{Meta-learning in Few-shot Learning}.
	Meta-learning, or learning to learn \cite{thrun2012learning}, is the technique of observing how different machine learning approaches perform on a wide range of tasks rather than batches of data points, and then starting from this experience, or meta-data, to learn new tasks much faster and obtain higher performance. In recent few-shot learning literature, more and more approaches follow the idea of meta-learning to alleviate over-fitting. Meta-LSTM \cite{ravi2016optimization} aims to learn efficient parameter updating rules for training neural network (learner) with a LSTM-based meta-learner. MAML \cite{finn2017model}, on the other hand, tries to learn a good model parameter initialization strategy that generalizes better to similar tasks. Similarly, Reptile \cite{nichol2018first} is an approximation of MAML that executes stochastic gradient descent with a first-order form. Meta-SGD \cite{li2017meta} goes further in meta-learning by arguing to learn the weights initialization, gradient update direction and learning rate within a single step. However, these approaches mentioned above suffer from the issue of fine-tuning. In contrast, after optimizing a meta-learner over a series of tasks, our approach solves unseen target tasks in an feed-forward manner without any further model updates and just optimize an episodic-wise distance metric efficiently for each task, thus avoiding gradient computation and serious overfitting.
	
	\noindent
	\textbf{Distance Metric Learning Approaches.} 
	Another category of approach focus on obtaining a generalizable embedding model to transform all samples into a common metric space where can perform simple classifiers such as nearest neighbor directly.  Matching Network \cite{vinyals2016matching} integrates metric learning with meta-learning for the first time by training a learnable nearest neighbor classifier with deep neural networks. Prototypical Network \cite{snell2017prototypical} utilizes class prototype representations to assign labels for query points and formulates the final loss function with euclidean distance directly. Later, this work is extended to semi-supervised few-shot scenario by \cite{ren2018meta}, unlabeled data are used to refine the class prototypes. Relation Network \cite{yang2018learning} trains an auxiliary network to compute the similarity score between each query and the support set, which is equivalent to further learn a non-linear metric. These approaches assume all samples are embedded into a task-agnostic metric space. Instead, our framework intends to emphasize the specificity among different tasks, that is, samples from different tasks should be embedded into a more discriminative task-specific space.
	
	\noindent
	\textbf{Task Adaptation Approaches.}
	The third family of approaches, which aims to mine the task adaptability with the meta-learning paradigm, is currently a hot research direction for few-shot learning. MT-Net \cite{lee2018gradient} proposes that the meta-learner learns on each layer’s activation space and the task-specific learner performs gradient descent on a subspace, which is more sensitive to task identity. TADAM \cite{oreshkin2018tadam} proposes a Task Encoding Network to produce scaling and shift vectors for each layer weights, leading to a task-dependent metric space. Our approach is slightly similar to these methods in the sense that we all focus on the idea of task adaptation. However, without imposing auxiliary network or applying complicated training strategy, our proposed framework formulates the adaptation procedure into a standard SDP problem under the transductive inference setting, which is more efficient and convenient.
	
	
	\noindent
	\textbf{Transductive Inference.} 
	Transductive inference \cite{vapnik1999overview} follows the setting of generalizing from the training set to the test set directly and avoiding the intermediate problem of estimating a function. In data-scarce scenario, it can significantly improve the performance over inductive methods. Reptile \cite{nichol2018first}, which implicitly shared information between all test samples via batch normalization \cite{ioffe2015batch}, is the first work of applying the transductive setting in few-shot learning. TPN \cite{Liu2018} models the transductive inference explicitly by learning a graph construction module, which propagates labels from labeled instances to unlabeled query points directly. Different from using example-wise parameters for the static Euclidean distance in \cite{Liu2018}, our approach explores the episodic-wise distance metric by integrating the small support set with the entire query set for transduction. 
	
	\section{Transductive Episodic-wise Adaptive Metric}
	
	In this section, we describe the definition of few-shot learning problem (FSL) and then introduce the \textit{Transductive Episodic-wise Adaptive Metric} (TEAM) in details.
	
	\subsection{Problem Formulation}
	
	Just like what is employed in various previous works \cite{ren2018meta, snell2017prototypical, vinyals2016matching, yang2018learning}, we organize the learning procedure into the form of episodic paradigm, which gradually collects meta-knowledge across a pool of source tasks and performs adaptation on target tasks quickly. Under this setting, the ultimate goal of our algorithm is to train models with a large labeled dataset $\mathcal{D}_{train}$, which is composed of a set of seen classes $\mathcal{C}_{train}$, and apply the classifiers on a novel testing set $\mathcal{D}_{test}$ with many unseen classes $\mathcal{C}_{test}$. Note that there are only a few labeled examples for each category in $\mathcal{C}_{test}$ and $\mathcal{C}_{test} \cap \mathcal{C}_{train} = \varnothing$. In order to mimic the few-shot test scenario during the training procedure and take full advantage of the large quantities of labeled $\mathcal{D}_{train}$, we reorganize all examples in $\mathcal{D}_{train}$ as a series of $N$-way $K$-shot tasks (or episodes). Concretely, $N$-way $K$-shot task is usually constructed by first selecting $N$ classes from $\mathcal{C}_{train}$ randomly and then generate a support set and a query set from the selected classes. The support set includes $K$ samples per classes, termed as $\mathcal{S}=\{(x_i, y_i)\}^{N \times K}_{i=1}$, and whilst the query set $\mathcal{Q}=\{(\hat{x}_1, \hat{y}_1), \dots, (\hat{x}_M, \hat{y}_M)\}$ contains different samples from the same label space with $\mathcal{S}$. In each episode, we train the learner with small labeled support set $\mathcal{S}$ and minimize the loss on the large query set $\mathcal{Q}$. After training episode by episode until convergence, the learned model can perform pretty well on novel few-shot tasks.  
	
	However, we argue that it is not ideal to apply the learned model to all target tasks directly without considering the specificity of them. Our approach proposes an episodic-wise metric construction module to transform the task-agnostic embeddings into a task-specific metric space, namely episodic adaptation. Moreover, in order to mitigate the data-scarce problem of support set and construct a more generalizable task-adaptive metric, we follow the paradigm of transductive inference and consider the query set as a whole for prediction, instead of one by one. 
	
	Specifically, our few-shot classification framework (see Fig. \ref{fig:framework}) is composed of three modules: (1) learning a task-agnostic feature extractor to embed raw inputs into a shared embedding space, this procedure includes a loss function to drive the parameters update and a novel task-level augmentation strategy to boost the generalization, (2) tailoring an episodic-wise adaptive metric for each task by solving a standard SDP problem efficiently and (3) performing a novel bi-directional similarity strategy in the task-specific space to assign label for each query. It is worth noting that the latter two modules utilize the setting of transduction. We describe the details of each module in the following sections.
	
	\subsection{Task-agnostic Feature Extractor}
	
	\noindent
	\textbf{Learning Embedding Function}. \ 
	Our approach first employs an embedding function $f_{\theta}$ to extract feature of an instance $x$, where $f_{\theta}(x)$ refers to the embedding of $x$ and $\theta$ indicates the parameters of deep model. Given the few-shot task sequence $\mathcal{T}$ from $\mathcal{D}_{train}$, we train the feature extractor $f_{\theta}$ episode by episode with minimizing the negative log-probability of true label for each sample via SGD as: 
	\begin{equation}  
	\underset{f_{\theta}}{\operatorname{argmin}} \ \ \mathop{\sum}_{ \mathcal{T}_i \backsim \mathcal{T}} \ \mathop{\sum}_{(x, y) \ \in \ \textit{TIM}(\mathcal{T}_i)} -log\ p(y \ \vert \ f_{\theta}(x, M_{\mathcal{T}_i}))
	\label{Eqn:1}
	\end{equation}
	where $\mathcal{T}_i$ is a few-shot episode sampled from task sequence randomly,  $\textit{TIM}(\cdot)$ is the novel task-level data augmentation operation and $M_{\mathcal{T}_i}$ stands for the episodic-wise adaptation metric of $\mathcal{T}_i$. Until convergence, the optimal embedding model $f_{\theta}^{\ast}$ is applied directly to unseen target tasks which are sampled from $\mathcal{D}_{test}$ instead of $\mathcal{D}_{train}$. 
	
	\noindent
	\textbf{Task Internal Mixing Augmentation Strategy}. Recently, some data augmentation techniques, such as flipping, rotation or distorting the inputs, follow the learning principle named Vicinal Risk Minimization (VRM) \cite{chapelle2001vicinal} to improve the generalization performance of deep neural networks. Inspired by \cite{inoue2018data, zhang2017mixup}, we further propose a task-level data augmentation technique which is termed as Task Internal Mixing (\textit{TIM}), performing the convex combination between all support samples in each task to synthesize new episodes. To be concrete, for each instance $(x_i, y_i)$ in a source task, we randomly select another sample $(x_j, y_j)$ from the same task and synthesize new training examples $(\tilde{x}, \tilde{y})$ as follows: 
	\begin{equation}
	\tilde{x} = \omega \cdot x_i + (1-\omega) \cdot x_j, \ \ \ \tilde{y} = y_i
	\label{Eqn:2}
	\end{equation}
	where $\omega \backsim U(l, \ h)$ and $ 0.5 \leqslant l < h \leqslant 1.0 $. Note $x_i$ and $x_j$ are just the raw input tensors instead of features. Then we handle each instance a few times with Eq.\ \eqref{Eqn:2} to form a virtual task (i.e.\ $ \textit{TIM}(\mathcal{T}_i) $ in Eq.\ \eqref{Eqn:1} ) for training. In essence, \textit{TIM} extends the source task distribution by incorporating the prior that if two samples are similar to each other in the original pixel space, then they are likely to be closer in the feature space. As such, $x_i$ and $\tilde{x}$ are more similar to each other than $x_j$ and $\tilde{x}$ in Eq.\ \eqref{Eqn:2} because of $\omega > 0.5$, which leads to the synthetic label $\tilde{y}$ should be $y_i$ instead of $y_j$.
	
	\subsection{Episodic-wise Adaptive Metric}
	
	\noindent
	\textbf{Distance with Metric $M_t$.} \
	Given two embeddings $x_i$ and $x_j$ in vector space $ \mathcal{V} \in \mathbb{R}^d $, we denote the distance between them with the metric $M_t$ as follows: 
	\begin{equation}
	d_{M_t}(x_i, x_j) = \sqrt{tr \langle M_t (x_i-x_j)(x_i-x_j)^\mathrm{T} \rangle}
	\label{Eqn:3}
	\end{equation}
	where $tr \langle \cdot \rangle$ stands for the trace operator and $M_t$ is a symmetric positive semi-definite matrix, which ensures that $d_{M_t}$ satisfies the properties of a pseudo-distance \cite{bellet2013survey}. In general, the matrix $M_t$ parameterizes a family of Mahalanobis distances in the vector space $ \mathcal{V} $. In particular, the $d_{M_t}$ in Eq.\ \eqref{Eqn:3} will degenerate into the popular Euclidean distance if we set $M_t=I$, that is, assuming all features are equally scaled and equally relevant \cite{torresani2007large}. Inspired by these observations, we proposed to explicitly construct episodic-wise adaptive metric by leveraging the specific information of each task, such as the pairwise constraints and the regularization prior. \\
	\
	\textbf{Pair-constrained Loss.} \
	Given a few-shot task, our goal is to minimize the mean of distances between all similar sample pairs (must-link constraints, denoted with $\mathcal{M}$) while keeping the mean of distances between all dissimilar sample pairs (cannot-link constraints, denoted with $\mathcal{C}$) larger than 1 at meanwhile. Motivated by the above idea, we formulate the \textit{Min-Max} principle as a convex optimization problem in Eq.\ \eqref{Eqn:4} by adopting the square distance in terms of its effectiveness and efficiency.
	\begin{equation}
	\begin{split}
	\mathop{\min}_{M_t \succeq 0} 
	\ \ \ & \frac{1}{\vert \mathcal{M} \vert }
	\ \ \ \mathop{\sum}\nolimits_{(x_i, x_j) \in \mathcal{M}} \ \ d_{M_t}^2(x_i, x_j) \\
	s.t. 
	\ \ \ & \ \frac{1}{\vert \mathcal{C} \vert } \ \
	\ \ \ \mathop{\sum}\nolimits_{(x_i, x_j) \in \ \mathcal{C} \ } \ \ d_{M_t}^2(x_i, x_j) \ \ge \ 1 \
	\label{Eqn:4}
	\end{split}
	\end{equation}
	Based on Eq.\ \eqref{Eqn:3} and the methods of lagrange multiplier, we rewrite the Eq.\ \eqref{Eqn:4} into a pair-constrained loss function as:
	\begin{equation}
	\mathcal{L}_C (M_t \ \vert \ \mathcal{\widetilde{M}}, \mathcal{\widetilde{C}}) = 
	\mathnormal{tr}(M_t \cdot \mathcal{\widetilde{M}}) - \lambda \cdot \mathnormal{tr}(M_t \cdot \mathcal{\widetilde{C}} \ )
	\label{Eqn:5}
	\end{equation}
	where $\lambda$ is the multiplier, $\mathcal{\widetilde{M}}$ and $\mathcal{\widetilde{C}}$ have the following form: 
	\begin{equation}
	\begin{split}
	& \mathcal{\widetilde{M}} = \frac{1}{\vert \mathcal{M} \vert } 
	\mathop{\sum}\nolimits_{(x_i, x_j) \in \mathcal{M}} (x_i-x_j)(x_i-x_j)^\mathrm{T} \
	\\
	& \ \mathcal{\widetilde{C}} \ = \ \frac{1}{\vert \ \mathcal{C} \ \vert }
	\mathop{\sum}\nolimits_{(x_i, x_j) \in \ \mathcal{C} \ } (x_i-x_j)(x_i-x_j)^\mathrm{T} \
	\label{Eqn:6}
	\end{split}
	\end{equation}
	Furthermore, given a $N$-way $K$-shot task with support set $\mathcal{S} = \{(x_i,\ y_i)\}_{i=1}^{NK}$ and its query set $\mathcal{Q}$, we first shrink the support set $\mathcal{S}$ to a prototype set $\mathcal{P}=\{p_c\}_{c=1}^{N}$  by: 	
	\begin{equation}
	p_c = \frac{1}{\vert \mathcal{S}_c \vert} \mathop{\sum}\nolimits_{(x_i, \ y_i) \in \mathcal{S}_c} x_i
	\label{Eqn:11}
	\end{equation}
	where $\mathcal{S}_c$ is the subset which contains the samples with the same label $c$ in $\mathcal{S}$. The we define the similar and dissimilar constraints with
	$\mathcal{M} = \{ x_i \in \mathcal{S}_c, x_j \in \mathcal{S}_c, i \ne j \} \cup 
	\{ x_i \in \mathcal{S}_c, x_j \in \mathcal{P}_c \} \cup 
	\{ x_i \in \mathcal{S}_c, x_j \in \mathcal{N}(x_i, k, \mathcal{Q}) \} $ and
	$\mathcal{C} = \{ x_i \in \mathcal{P}_c, x_j \in \mathcal{P}_{c'}, c \ne c' \} \cup \{ x_i \in \mathcal{P}_c, x_j \in \mathcal{P}_T \}$
	where $\mathcal{P}_c$ is a prototype with label $c$ in $\mathcal{P}$, $\mathcal{N}(x_i, k, \mathcal{Q})$ is a set of $k$ nearest neighbors of $x_i$ in the query set $\mathcal{Q}$ and $\mathcal{P}_T$ is the prototype set from the seen classes $\mathcal{C}_{train}$. 
	
	\noindent
	\textbf{Regularization Loss.} \
	Without imposing any restrictions or prior information into the Eq.\ \eqref{Eqn:5}, we have $d^2$ parameters to be optimized for each task due to $M_t \in \mathbb{R}^{d \times d}$, while we can only construct a few pairwise constraints in few-shot scenario. From the perspective of machine learning theory, this inconsistency will lead to severe over-fitting of our models. To this end, we propose the second principle to regularize the episodic-wise metric $M_t$ to be close to a given metric $M_0$, which is associated with the prior across all few-shot tasks. Specifically, we try to minimize the Bregman divergence 
	$D_{\Phi}(M_t \Vert M_0) = \Phi(M_t) - \Phi(M_0) - \langle \nabla \Phi(M_0), M_t-M_0 \rangle$ 
	between $M_t$ and $M_0$ with the log-determinant function $\Phi(M) = -log \ det(M)$, which is a strict convex, continuously differentiable divergence function. Then we formulate the regularization loss function 
	$\mathcal{L}_R (M_t \ \vert M_0) $ as: 
	\begin{equation}
	\mathcal{L}_R (M_t \ \vert \ M_0) = \mathnormal{tr} \langle M_0^{-1} M_t \rangle - \mathnormal{log \ det}(M_t)
	\label{Eqn:7}
	\end{equation}
	where $tr \langle \cdot \rangle$ means the trace operator on matrix and Eq.\ \eqref{Eqn:7} ignores the constant term regarding ${\mathcal{M}_0}$. More precisely, with the information-theory, optimizing $\mathcal{L}_R $ is equivalent to minimizing the KL divergence between two multivariate Gaussian distributions parameterized by $M_t$ and $M_0$.
	\\
	\textbf{Episodic-wise Adaptive Metric.} \ 
	By integrating two principles mentioned above, we formulate a novel episodic-wise adaptive metric (\textit{EAM}) loss function for each task as:
	\begin{equation}
	\mathop{\min}_{M_t \succeq 0} \ \ 
	\mathcal{L}_R (M_t \ \vert \ M_0) +  \gamma \cdot 
	\mathcal{L}_C (M_t \ \vert \ \mathcal{\widetilde{M}}, \ \mathcal{\widetilde{C}} \ )
	\label{Eqn:8}
	\end{equation}
	where $\mathcal{L}_C$ and $\mathcal{L}_R$ are defined in Eq.\ \eqref{Eqn:5} and Eq.\ \eqref{Eqn:7} respectively, $\gamma$ is a positive trade-off parameter. In general, minimizing Eq.\ \eqref{Eqn:8} with SGD-based optimizers or other existing convex optimization solvers \cite{sturm1999using} will produce a local optimal solution for each task. However, due to the high time complexity of SDP solvers and requiring too many iterations with SGD-based optimizers, this sub-optimization procedure within each task will lead to an inefficient learning process. Here, we propose a faster and more efficient approach to construct the episodic-wise metric for each task. Concretely, we first reformulate the Eq.\ \eqref{Eqn:8} as follows:
	\begin{equation}
	\mathop{\min}_{M_t \succeq 0}
	\mathnormal{tr} \langle M_t \cdot (\mathcal{M}_{0}^{-1} + \gamma \cdot \mathcal{\widetilde{M}} - \gamma \lambda \cdot \mathcal{\widetilde{C}} ) \rangle - \mathnormal{log \ det} (M_t)
	\label{Eqn:7-1}
	\end{equation}
	Based on the Lemma \eqref{lemma:1}, we get the optimal solution $M_t^*$ by letting $\mathcal{Y}=\mathcal{M}_{0}^{-1} + \gamma \cdot \mathcal{\widetilde{M}} - \gamma \lambda \cdot \mathcal{\widetilde{C}}$ and assuming $\mathcal{Y} \succ 0$,  
	\begin{equation}
	M_t^*= (\mathcal{M}_{0}^{-1} + \gamma \cdot \mathcal{\widetilde{M}} - \gamma \lambda \cdot 
	\mathcal{\widetilde{C}}\ )^{-1}
	\label{Eqn:9}
	\end{equation}
	And the assumption of $\mathcal{Y} \succ 0$ can always hold if we pick up a positive definite matrix as prior metric $M_0$. In addition to considering the pairwise constraints and regularization prior, we further introduce the feature correlation information into the final metric $M_t^\dagger$ with the task covariance matrix 
	$\Sigma_t = \frac{1}{n-1} \mathop{\sum}_{i=1}^{n} \mathcal(X_i-\bar{X}) \mathcal(X_i-\bar{X})^\mathnormal{T}$, which results in:
	\begin{equation}
	M_t^\dagger = (\mathcal{M}_{0}^{-1} + \gamma \cdot \mathcal{\widetilde{M}} - \gamma \lambda \cdot 
	\mathcal{\widetilde{C}})^{-1} + \alpha \cdot \Sigma_t
	\label{Eqn:10}
	\end{equation}
	where $\alpha$, $\lambda$, $\gamma$ are positive trade-off parameters and $\mathcal{\widetilde{M}}$, $\mathcal{\widetilde{C}}$ are from Eq. \eqref{Eqn:6}. Moreover, with the insights of transduction, we calculate the task covariance matrix $\Sigma_t$ with both the support set and query set in each episode. Obviously, the Eq.\ \eqref{Eqn:10} only involves simple matrix operations, such as inversion and transpose, which is more efficient than SGD-based optimizers and naive SDP solvers. In addition, as a symmetric positive definite matrix, another insights into the nature of the learned episodic-wise metric is an adaptive linear projection layer by expressing $M_t^\dagger$ as ${L_t}^\mathrm{T}L_t$, then the $d_{M_t}(x_i, x_j)$ in Eq.\eqref{Eqn:3} can be formulated as $\sqrt{tr \langle (L_t{x_i}-L_t{x_j})^\mathrm{T}(L_t{x_i}-L_t{x_j}) \rangle }$, where $L_t \in \mathbb{R}^{r \times d}$ is a task-specific transformation matrix.
	\vspace{-0.2cm}
	\begin{lemma}
		Let $\mathcal{X}, \mathcal{Y}$ be two symmetric positive-define matrices of the same size, then the function
		$f(\mathcal{X})=\mathnormal{tr}(\mathcal{XY}) - log \ det(\mathcal{X})$
		is minimized uniquely by:
		$\mathcal{X}^*=\mathcal{Y}^{-1}.$ 
		\label{lemma:1}
	\end{lemma}
	\vspace{-0.2cm}
	\textit{Proof.} See supplementary material for more details.
	\vspace{-0.1cm} \
	
	\subsection{Bi-directional Similarity}
	Assuming all samples have been transformed into a task-specific embedding space with the learned feature extractor $f_{\phi}$ and the episodic-wise adaptive metric $M_t^\dagger$ in Eq.\ \eqref{Eqn:10}, we then perform a novel bi-directional similarity strategy (\textit{Bi-SIM}) to calculate the probability that each query belongs to each category. In detail, after shrinking the the support set $\mathcal{S}$ to a prototype set with Eq.\eqref{Eqn:11}, we formulate the \textit{positive-direction} similarity $s_{i \to c}$ between query $x_i$ and each prototype $p_c$ with the softmax function:
	\begin{equation}
	s_{i \to c} = \frac{\exp({-d_{M_t}(x_i, \ p_c)})} {\sum_{c'}\exp({-d_{M_t}(x_i, \ p_{c'})})}
	\label{Eqn:12}
	\end{equation}
	Most previous methods used this similarity as the final probability of the query $x_i$ belonging to each category. However, taking the entire query set into account with transductive inference, we further compute the probability of prototype $p_c$ belonging to each query $x_i$ with the following equation:
	\begin{align}
	s_{c \to i} = \frac{\exp({-d_{M_t}(p_c, \ x_i)})}{\sum_{i'}\exp({-d_{M_t}(p_c, \ x_{i'})})}
	\label{Eqn:13}
	\end{align} 
	We termed $s_{c \to i}$ as the \textit{negative-direction} similarity which can be interpreted as an attention-based weight of the prototype $p_c$ over the whole query set $\mathcal{Q}$. At last, we perform the product of $s_{i \to c}$ and $s_{c \to i}$ as the final bi-directional similarity (denoted with \textit{Bi-Sim}) between the query $x_i$ and the prototype $p_c$, i.e.\ $s_{i \leftrightarrow c} = s_{c \to i} \cdot s_{i \to c}$. Essentially, the basic idea behind the \textit{Bi-Sim} strategy is that if a query is similar to one prototype and the prototype is also similar to the query, then we argue that they are more matching with each other. Without increasing any computational burden or requiring any human interaction, our proposed strategy calculates more robust similarity efficiently. 
	
	\section{Experiments}
	In this section, we detail our experimental setting and compare TEAM with state-of-the-art approaches on three challenging datasets, i.e.\ \textit{mini}ImageNet \cite{vinyals2016matching}, Cifar-100 \cite{krizhevsky2009learning} and CUB \cite{wah2011caltech}, which are widely used as few-shot classification benchmarks in the literature. 
	
	\subsection{Datasets}
	
	\noindent
	\textbf{\textit{mini}ImageNet.} \
	The \textit{mini}ImageNet dataset is the most popular benchmark in few-shot learning community, which is proposed by \cite{vinyals2016matching} originally. This dataset is composed of 100 classes selected from ImageNet  \cite{krizhevsky2012imagenet} randomly, and each class has 600 images, which are resized to $84 \times 84$ pixels for fast training and inference. Note that we follow the setup provided by \cite{ravi2016optimization} which splits the total 100 classes into 64 classes, 16 classes and 20 classes for training, validation and evaluation respectively. The validation set is only used for tracking model generalization in all experiments.
	
	\noindent
	\textbf{Cifar-100.} \
	The Cifar-100 \cite{krizhevsky2009learning} is a simple dataset for image classification and consists of 100 categories, each having 600 RGB images ($32 \times 32$). We further split the whole dataset into 64 classes as seen categories for training, 16 and 20 classes for validation and testing respectively \cite{zhou2018deep}. Compared with \textit{mini}ImageNet, Cifar-100 keeps the simplicity of dataset and decreases the inference complexity.
	
	\noindent
	\textbf{CUB.} \
	CUB \cite{wah2011caltech}, which is a benchmark dataset for fine-grain classification initially, is composed of 11788 images over 200 birds classes. Follow the same partition as \cite{hilliard2018few}, we use 100 classes for training, and another two 50 classes as unseen classes for validation and evaluation. And all images are cropped with the provided bounding box \cite{triantafillou2017few}.
	
	\subsection{Experimental Settings}
	
	\noindent
	\textbf{Backbone Networks.} \
	For fair and comprehensive comparison with previous baselines, we employ two backbone networks as our embedding function. 1) A four-layers convolution network (ConvNet) and 2) A standard deep residual network (ResNet-18) are widely adopted in the few-shot learning literature. Specifically, the ConvNet contains 4 repeated convolutional blocks, where each block is composed of a convolution layer with 64 filters ($3 \times 3$ kernel), a batch normalization layer \cite{ioffe2015batch}, a ReLU non-linearity and a max-pooling layer with size $2$. In addition, we empirically add a global average pooling layer as last for accelerating the convergence of the model and reducing the dimensionality of the features. All inputs are resized to $84 \times 84 \times 3$ uniformly and the final output dimension is 256 for each image accordingly. For ResNet, we utilize the standard architecture proposed by \cite{he2016deep} and remove the last fc-layer for reducing parameters. Furthermore, all inputs are resized to $224 \times 224 \times 3$ like many previous works. After the last average pooling layer, it leads to a 512 vector for each image.
	
	\noindent
	\textbf{Training Strategy.} \
	All backbone networks are optimized via SGD by Adam \cite{kingma2014adam} end-to-end on DGX-1. Follow the strategy in \cite{qiao2017few, rusu2018meta}, we pre-train the ConvNet to classify all seen classes and utilizing the optimal weights for model initialization, and we train ResNet from scratch for simplicity. Moreover, We perform \textit{TIM} strategy in all experiments and set $l=0.5$ and $h=1.0$ for $\mathcal{U}(l,\ h)$. Inspired by \cite{inoue2018data}, we start \textit{TIM} strategy after 5000 episodes and intermittently disable it during training procedure, that is, performing task mixing for $Y$ episodes and then close it for the next $Z$ episodes. We empirically set $Y=4$ and $Z=1$ in our experiments. We decay the learning rate half every 10000 episodes and set the patience of early stopping as 20000.
	
	\noindent
	\textbf{Parameter Setup in $M_t^{\dagger}$.} \
	In Eq.\ \eqref{Eqn:8}, we set the trade-off parameters as $\alpha=2$, $\gamma=0.2$, $\lambda=0.01$ and the prior metric as $M_0=I$ in all experiments. In general, the choice of $M_0$ is not fixed and has an important influence on generalization of the learned episodic-wise metric. However, based on the following two observation, we argue the identity matrix is a quite natural choice for $M_0$. Firstly, learning from the Euclidean distance provides the most unbiased prior across all few-shot tasks, that is, assuming all features from the task-agnostic embedding space are equally scaled and equally relevant. Secondly, we observe that the optimal adaptive metric for each few-shot task is close to the identity matrix, which has been illustrated in Fig. \ref{fig:sparsity} of the paper. Please zoom in the Fig. \ref{fig:sparsity} or refer appendix for more details.
	
	\subsection{Few-shot Learning Results}
	
	To verify the effectiveness of our approach for few-shot classification, we compare the proposed TEAM framework with our re-implemented baseline (ProtoNet \cite{snell2017prototypical}) and many state-of-the-art methods in various setting on three benchmark datasets (\textit{mini}ImageNet, Cifar-100 and CUB). For fair comparison with previous works, we focus on two popular few-shot learning settings, namely 5-way 1-shot and 5-way 5-shot tasks, which both contain 15 queries per episode for validation. In addition to the above setup, we also experimente with the transductive setting in all datasets, where the model utilizes the entire query set in each task. Specifically, we consider two types of transduction in our experiments. 1) Transductive batch normalization \cite{finn2017model, nichol2018first}, which shares information between all test examples via batch normalization layer, denoted with \textit{BN} in all tables and 2) explicit transduction, which is first introduced into the few-shot learning by \cite{Liu2018}. Moreover, to make the evaluation more convincing, we report the final mean accuracy over 1000 test trails for all experiments and present 95\% confidence intervals of all results in our supplementary material. Please refer our appendix for more complete results. 
	
	\noindent
	\textbf{Results on \textit{mini}ImageNet.} \
	Experimental results on \textit{mini}ImageNet are shown in Table \ref{tab:miniImageNet}, where we can see that our model achieves state-of-the-art performance with ConvNet backbone and competitive results with ResNet architecture. We re-implement ProtoNet as our baseline with the simple pre-train strategy proposed by \cite{qiao2017few}, and achieve better performance than previously reported ones in \cite{snell2017prototypical}. Taking ConvNet as an example, we get 51.68\% and 68.71\% for 5-way 1-shot and 5-way 5-shot respectively, which are slightly better than 49.42\% and 68.20\% in \cite{snell2017prototypical}. After applying the TEAM framework on the baseline, the performance has been further improved significantly. For example, the absolute promotion of TEAM over published state-of-the-art is 1.06\% for 1-shot and 2.18\% for 5-shot, over our baseline is 4.89\% and 3.33\% respectively. Note that the comparisons with PFA \cite{qiao2017few}, LEO\cite{rusu2018meta} and TADAM \cite{oreshkin2018tadam} are a bit unfair since we train the  TEAM(ResNet) without any pre-train weights or including a classification objective, however, our model still achieves the best performance on 1-shot task.
	
	\renewcommand{\arraystretch}{1.15} 
	\begin{table}[tp]
		\centering
		\fontsize{7.8}{8}\selectfont
		\caption{Few-shot classification accuracy on \textit{mini}ImageNet. All results are averaged over 1000 test tasks which are randomly selected from the testing set. Tran: different type of transduction.}
		\label{tab:miniImageNet}
		~\\[-1mm]
		\scalebox{0.97} 
		{
			\begin{tabular}{lccccc}
				\toprule[1.1pt]
				\multirow{2}{*}{Model} &\multirow{2}{*}{Tran.} &\multicolumn{2}{c}{5-Way 1-Shot} &\multicolumn{2}{c}{5-Way 5-Shot} \\
				\cmidrule(lr){3-6} & &ConvNet &ResNet &ConvNet &ResNet \\
				\midrule
				MatchNet \cite{vinyals2016matching}   & No  & 43.56 &   -   & 55.31 &   -   \\
				MAML \cite{finn2017model}             & BN  & 48.70 &   -   & 63.10 &   -   \\
				MAML+ \cite{Liu2018}                  & Yes & 50.83 &   -   & 66.19 &   -   \\
				Reptile \cite{nichol2018first}        & BN  & 49.97 &   -   & 65.99 &   -   \\
				ProtoNet \cite{snell2017prototypical} & No  & 49.42 &   -   & 68.20 &   -   \\
				GNN \cite{garcia2017few}              & No  & 50.33 &   -   & 64.02 &   -   \\
				RelationNet \cite{yang2018learning}   & BN  & 50.44 &   -   & 65.32 &   -   \\
				PFA \cite{qiao2017few}                & No  & 54.53 & 59.60 & 67.87 & 73.74 \\
				TADAM \cite{oreshkin2018tadam}        & No  &   -   & 58.50 &   -   & \textbf{76.70} \\
				adaResNet \cite{munkhdalai2018rapid}  & No  &   -   & 56.88 &   -   & 71.94 \\
				LEO \cite{rusu2018meta}               & No  &   -   & 60.06 &   -   & 75.72 \\
				TPN \cite{Liu2018}                    & Yes & 55.51 & 59.46 & 69.86 & 75.65 \\
				\hline \\ [-2.5mm]
				Baseline (Ours)                       & No  & 51.68 & 55.25 & 68.71 & 70.58\\
				TEAM (Ours)                           & Yes & \textbf{56.57} & \textbf{60.07} & \textbf{72.04} & 75.90 \\
				\bottomrule[1pt]
			\end{tabular}
		}
		\vspace{-0.5cm} 
	\end{table}
	
	\noindent
	\textbf{Results on Cifar-100.} \
	Next we turn to the rich experiments evaluated on the Cifar-100, and all results are shown in Table \ref{tab:cifar} for comparison in detail. Note that all results of MatchNet \cite{vinyals2016matching}, MAML \cite{finn2017model} and DEML \cite{zhou2018deep} in Table \ref{tab:cifar} refer to the reported performance in \cite{zhou2018deep}. Compared with our baseline, whose accuracy is slightly higher than previous points, we notice that our TEAM(ConvNet) increases by 6.24 \% on 1-shot tasks and 2.65 \% on 5-shot tasks, which demonstrates the effectiveness of our approach. 
	
	\noindent
	\textbf{Results on CUB.} \
	The CUB dataset \cite{wah2011caltech} is proposed for fine-grain recognition initially and also widely utilized in current few-shot classification literature. From Table \ref{tab:cub}, we observe that the performance of our baseline is far better than the previous ProtoNet \cite{snell2017prototypical}, that is because we preprocess all images with the provided bounding boxes \cite{triantafillou2017few} to reduce the impact of background on final performance. Comparing the TEAM with our re-implemented baseline, both ConvNet and ResNet backbones achieve outstanding performance over 1-shot and 5-shot tasks.
	
	\noindent
	\textbf{Further Analysis.} \
	All results which are summarized in the Table \ref{tab:miniImageNet} - \ref{tab:cub} indicate that our approach can consistently improve the few-shot learning performance on different datasets. This confirms that, under the setting of transductive inference, our model can efficiently tailor an episodic-wise adaptive metric for each task and perform a suitable similarity between all samples. Furthermore, we notice that the performance promotion of our approach in 1-shot scenario is more significant than that in 5-shot. This observation agrees with the nature of transduction \cite{joachims1999transductive, Liu2018}, where more training data are available, the less performance improvement will be. With regards to this, we then perform 5-way $k$-shot (k=1, 3, 5, 7, 9) experiments on \textit{mini}-ImageNet and all results are shown in Table. \ref{tab:query}. As the number of shots increases, we notice that our TEAM consistently outperforms our baseline with a large margin, but the performance improvement from TEAM decreases slightly, which further verifies the above analysis about transductive inference.

	
	
	\renewcommand{\arraystretch}{1.15} 
	\begin{table}[tp]
		\centering
		\fontsize{7.8}{8}\selectfont
		\caption{Few-shot classification performance on Cifar-100.}
		\label{tab:cifar}
		~\\[-1mm]
		\scalebox{0.97} 
		{
			\begin{tabular}{lccccc}
				\toprule[1pt]
				\multirow{2}{*}{Model} &\multirow{2}{*}{Tran.} &\multicolumn{2}{c}{5-Way 1-Shot} &\multicolumn{2}{c}{5-Way 5-Shot} \\
				\cmidrule(lr){3-6} & &ConvNet &ResNet &ConvNet &ResNet \\
				\midrule
				MatchNet \cite{vinyals2016matching}   & No  & 50.53 &   -   & 60.30 &   -   \\
				MAML \cite{finn2017model}             & BN  & 49.28 &   -   & 58.30 &   -   \\
				ProtoNet \cite{snell2017prototypical} & No  & 56.66 &   -   & 76.29 &   -   \\
				DEML \cite{hariharan2017low}          & No  &   -   & 61.62 &   -   & 77.94 \\
				\hline \\ [-2.5mm]
				Baseline (Ours)                       & No  & 57.83 & 66.30 & 76.40 & 80.46 \\
				TEAM (Ours)                           & Yes & \textbf{64.07} & \textbf{70.43} & \textbf{79.05} & \textbf{81.25} \\
				\bottomrule[1pt]
			\end{tabular}
		}
		\vspace{-0.2cm}
	\end{table}
	
	\renewcommand{\arraystretch}{1.15} 
	\begin{table}[tp]
		\centering
		\fontsize{7.8}{8}\selectfont
		\caption{Few-shot classification performance on CUB.}
		\label{tab:cub}
		~\\[-1mm]
		\scalebox{0.97} 
		{
			\begin{tabular}{lccccc}
				\toprule[1pt]
				\multirow{2}{*}{Model} &\multirow{2}{*}{Tran.} &\multicolumn{2}{c}{5-Way 1-Shot} &\multicolumn{2}{c}{5-Way 5-Shot} \\
				\cmidrule(lr){3-6}                    &     &ConvNet&ResNet &ConvNet&ResNet \\
				\midrule
				MatchNet \cite{vinyals2016matching}   & No  & 56.53 &   -   & 63.54 &   -   \\
				MAML \cite{finn2017model}             & BN  & 50.45 &   -   & 59.60 &   -   \\
				ProtoNet \cite{snell2017prototypical} & No  & 58.43 &   -   & 75.22 &   -   \\
				RelationNet \cite{yang2018learning}   & BN  & 62.45 &   -   & 76.11 &   -   \\
				DEML \cite{hariharan2017low}          & No  &   -   & 66.95 &   -   & 77.11 \\
				TriNet \cite{chen2018semantic}        & No  &   -   & 69.61 &   -   & 84.10 \\
				\hline \\ [-2.5mm]
				Baseline (Ours)                       & No  & 69.39 & 74.55 & 82.78 & 85.98 \\
				TEAM (Ours)                           & Yes & \textbf{75.71} & \textbf{80.16} & \textbf{86.04} & \textbf{87.17} \\
				\bottomrule[1pt]
			\end{tabular}
		}
		\vspace{-0.4cm}
	\end{table}
	
	\subsection{Ablation Study}
	
	\noindent
	\textbf{Effectiveness of Different Modules.} \
	According to the previous analysis, the proposed TEAM framework is far superior to our baseline and becomes the new state-of-the-art approach in few-shot classification literature. As a necessary step to the ablation study, we first analyze how much each module (\textit{TIM}, \textit{EAM} and \textit{Bi-Sim}) contributes to the ultimate performance. All results are shown in Table \ref{tab:ablation} in great details. Note that our baseline (ProtoNet) uses the ConvNet as backbone network and achieves higher performance on all three datasets (see the second row in Table \ref{tab:ablation}) than previous performance because of the pre-train weights initialization. Furthermore, we perform various setting of TEAM framework as follows. 1) TEAM$^{\ddagger}$ adds the \textit{TIM} strategy into our baseline, 2) TEAM$^{\dagger}$ utilizes the \textit{TIM} strategy and episodic-wise adaptive metric (\textit{EAM}) simultaneously and 3) TEAM combines all three modules together to get the ultimate performance. By comparing the second and third rows in Table \ref{tab:ablation}, we observe that the \textit{TIM} strategy can consistently improve the performance of all few-shot tasks. Then we further compare the TEAM$^{\ddagger}$ and TEAM$^{\dagger}$, where the only difference between them is whether using \textit{EAM} module. Taking 1-shot task of \textit{mini}ImageNet as an example, TEAM$^{\dagger}$ is 2.38\% higher than TEAM$^{\ddagger}$, which demonstrates that it is feasible to construct an episodic-wise adaptive metric for each task in few-shot learning. And the last row of Table \ref{tab:ablation} further shows the effectiveness of our entire framework.
	
	\renewcommand{\arraystretch}{1.15} 
	\begin{table}[tp]
		\centering
		\fontsize{7.8}{8}\selectfont
		\caption{Few-shot classification performance for ablation study. Proto (Ours): the baseline. TEAM$^{\ddagger}$: baseline+\textit{TIM}. TEAM$^{\dagger}$: baseline+\textit{TIM}+\textit{EAM}. TEAM: baseline+\textit{TIM}+\textit{EAM}+\textit{Bi-SIM}. }
		\label{tab:ablation}
		~\\[-1mm]
		\scalebox{0.97} 
		{
			\begin{tabular}{lcccccc}
				\toprule[1pt]
				\multirow{2}{*}{Model} &\multicolumn{2}{c}{\textit{mini}ImageNet} &\multicolumn{2}{c}{Cifar-100} &\multicolumn{2}{c}{CUB} \\
				\cmidrule(lr){2-7} &1-shot &5-shot &1-shot &5-shot &1-shot &5-shot \\
				\midrule
				Proto \cite{vinyals2016matching}   & 49.42 & 68.20 & 56.66 & 76.29 & 58.43 & 75.22 \\
				\hline\\[-2.5mm]
				Proto (Ours)                       & 51.68 & 68.71 & 57.83 & 76.40 & 69.39 & 82.78 \\
				TEAM$^{\ddagger}$                  & 52.97 & 70.45 & 59.56 & 77.65 & 70.27 & 84.68 \\
				TEAM$^{\dagger}$                   & 55.35 & 71.59 & 62.76 & 78.80 & 75.06 & \textbf{86.06} \\ 
				TEAM                               & \textbf{56.57} & \textbf{72.04} & \textbf{64.07} & \textbf{79.05} & \textbf{75.71} & 86.04 \\
				\bottomrule[1pt]
			\end{tabular}
		}
		\vspace{-0.25cm}
	\end{table}
	
	\renewcommand{\arraystretch}{1.15} 
	\begin{table}[tp]
		\centering
		\fontsize{8}{8}\selectfont
		\caption{\textit{semi}-Supervised comparison on \textit{mini}ImageNet}
		\label{tab:semi}
		~\\[-1mm]
		\scalebox{1.1} 
		{
			\begin{tabular}{lcc}
				\toprule[1pt]
				Methods                                          & 5-way 1-shot     & 5-way 5-shot     \\
				\midrule
				Soft \textit{k}-Means \cite{ren2018meta}         & $ 50.09\pm0.45 $ & $ 64.59\pm0.28 $ \\
				Soft \textit{k}-Means+Cluster \cite{ren2018meta} & $ 49.03\pm0.24 $ & $ 63.08\pm0.18 $ \\
				Masked Soft \textit{k}-Means \cite{ren2018meta}  & $ 50.41\pm0.31 $ & $ 64.39\pm0.24 $ \\
				TPN-semi \cite{Liu2018}                          & $ 52.78\pm0.27 $ & $ 66.42\pm0.21 $ \\
				\hline \\ [-2.5mm] 
				TEAM-semi (Ours)                                 & \textbf{54.81} $\pm$ \textbf{0.59}  & \textbf{68.92} $\pm$ \textbf{0.38} \\
				\bottomrule[1pt]
			\end{tabular}
		}
		\vspace{-0.5cm}
	\end{table}
	
	\noindent
	\textbf{Comparison with \textit{semi}-Supervised Few-shot Learning.} \
	From the perspective of unlabeled data, transductive inference is a special case of semi-supervised learning, that is, the former one directly uses test set as unlabeled data and the latter uses more auxiliary unlabeled data. As such, we propose a semi-supervised version of the TEAM framework, namely TEAM-semi, to compare it with other semi-supervised few-shot approaches. Specifically, following the labeled/unlabeled data split in \cite{ren2018meta}, we use 40\% and 60\% in each class as labeled and unlabeled data respectively. Note that the support/query examples in each task are both randomly sampled from the labeled set only for fair comparison. All results, which are averaged over 10 random labeled/unlabeled partition of the training set, are reported in Table \ref{tab:semi} in details. Compared with the previous state-of-the-art approaches TPN \cite{Liu2018}, our TEAM-semi framework increases by 2.03\% and 2.50\% for 1-shot/5-shot respectively, which verifies its ability to handle both supervised and \textit{semi}-supervised few-shot classification.
	
\begin{figure}[t]
		\begin{center}
			\includegraphics[width=0.9\linewidth]{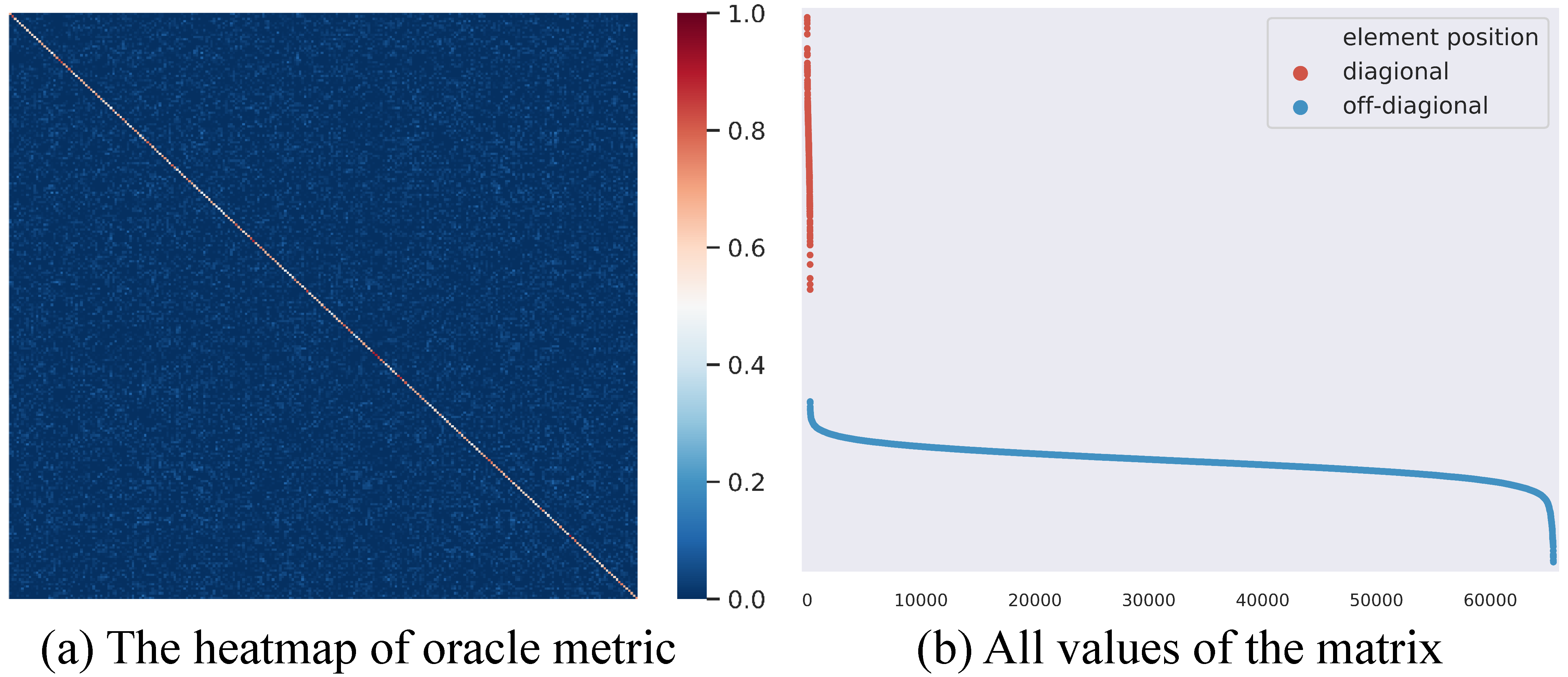}
		\end{center}
		\vspace{-0.25cm}
		\caption{This figure illustrates the sparsity nature of the metric in few-shot learning. Left: The heatmap of the oracle metric (Please zoom in it for more details). Right: The values distribution in different position of the matrix (sorted by descending order).}
		\label{fig:sparsity}
		\vspace{-0.2cm}
	\end{figure}

	\noindent
	\textbf{Sparsity Nature of Episodic-wise Adaptive Metric.} \
	In this section we explore the sparsity nature of the episodic-wise adaptive metric in few-shot learning. Take a 5-way 5-shot task as an example, we set 15-queries in each class and exploit the classic LMNN algorithm \cite{weinberger2006distance} with all support and query samples to optimize an oracle metric, which ensures all examples in this task can be completely distinguished. Then we scale all elements of the metric into the region [0, 1] and visualize its heatmap in Fig.\ \ref{fig:sparsity} (left). We observe that diagonal elements always maintain larger values (close to red) than off-diagonal elements (close to blue). After reorganizing all values with numerical descending order in Fig.\ \ref{fig:sparsity} (right), we further notice that there is a large value gap between the diagonal elements and off-diagonal elements. These practical observations indicate that, due to the low-data setup, we cannot have enough prior to find accurate correlations between all dimensions, except strong self-correlation in diagonal, which leads to the sparsity nature of episodic-wise adaptive metric. Moreover, from this practical viewpoint, we further verify that it is reasonable to set the identity matrix as prior metric $M_0$ in Eq.\ \eqref{Eqn:10}.
	
	
	\renewcommand{\arraystretch}{1.15}
	\begin{table}[tp]
		\centering
		\fontsize{8}{8}\selectfont
		\caption{5-way performance with various training/testing shots.}
		\label{tab:query}
		~\\[-1mm]
		\scalebox{1.03} 
		{
			\begin{tabular}{lccccc}
				\toprule[1pt]
				Methods          & 1-shot & 3-shot & 5-shot & 7-shot & 9-shot \\
				\midrule
				Baseline (Ours)  & 51.68  & 63.87  & 68.71  & 71.28  & 73.35  \\
				TEAM (Ours)      & 56.57  & 67.64  & 72.04  & 73.47  & 75.04  \\
				\hline \\ [-1.65 mm] 
				Accuracy (+)     &  4.89  &  3.77  &  3.33  &  2.19  &  1.69  \\
				\bottomrule[1pt]
			\end{tabular}
		}
		\vspace{-0.5cm}
	\end{table}
	
	\section{Conclusions}
	We have proposed \textbf{T}ransductive \textbf{E}pisodic-wise \textbf{A}daptive \textbf{M}etric (TEAM) for few-shot learning, which is a simple and efficient framework based on meta-learning. It not only learns a shared embedding model across all tasks end-to-end but also further tailors an episodic-wise metric by taking more distinctive information within each task into account. Moreover, with using the entire query set at once for inference, we leverage a bi-directional similarity strategy for extracting more robust relationship between queries and prototypes. Our TEAM achieves the state-of-the-art performance on three few-shot benchmark datasets and is easily extended to \textit{semi}-supervised version. The extensions of TEAM on other few-shot approaches could be future work.
	\\[6pt]
	\noindent
	\textbf{Acknowledgement.}
	This work is partially supported by grants from the National Key R\&D Program of China under grant 2017YFB1002400, the National Natural Science Foundation of China under contract No.U1611461, No.61825101 and No.61672072, also supported by grants from NVIDIA and the NVIDIA DGX-1 AI Supercomputer, and Beijing Nova Program (Z181100006218063). 

{\small
\bibliographystyle{ieee_fullname}
\bibliography{egbib}

\begin{thebibliography}{10}\itemsep=-1pt

\bibitem{bellet2013survey}
Aur{\'e}lien Bellet, Amaury Habrard, and Marc Sebban.
\newblock A survey on metric learning for feature vectors and structured data.
\newblock {\em arXiv:1306.6709}, 2013.

\bibitem{boyd2004convex}
Stephen Boyd and Lieven Vandenberghe.
\newblock {\em Convex optimization}.
\newblock Cambridge university press, 2004.

\bibitem{chapelle2001vicinal}
Olivier Chapelle, Jason Weston, L{\'e}on Bottou, and Vladimir Vapnik.
\newblock Vicinal risk minimization.
\newblock In {\em NIPS}, pages 416--422, 2001.

\bibitem{chen2018semantic}
Zitian Chen, Yanwei Fu, Yinda Zhang, Yu-Gang Jiang, Xiangyang Xue, and Leonid
  Sigal.
\newblock Semantic feature augmentation in few-shot learning.
\newblock {\em arXiv:1804.05298}, 2018.

\bibitem{finn2017model}
Chelsea Finn, Pieter Abbeel, and Sergey Levine.
\newblock Model-agnostic meta-learning for fast adaptation of deep networks.
\newblock In {\em ICML}, 2017.

\bibitem{garcia2017few}
Victor Garcia and Joan Bruna.
\newblock Few-shot learning with graph neural networks.
\newblock In {\em ICLR}, 2017.

\bibitem{hariharan2017low}
Bharath Hariharan and Ross~B Girshick.
\newblock Low-shot visual recognition by shrinking and hallucinating features.
\newblock In {\em ICCV}, pages 3037--3046, 2017.

\bibitem{he2016deep}
Kaiming He, Xiangyu Zhang, Shaoqing Ren, and Jian Sun.
\newblock Deep residual learning for image recognition.
\newblock In {\em CVPR}, pages 770--778, 2016.

\bibitem{hilliard2018few}
Nathan Hilliard, Lawrence Phillips, Scott Howland, Art{\"e}m Yankov, Courtney~D
  Corley, and Nathan~O Hodas.
\newblock Few-shot learning with metric-agnostic conditional embeddings.
\newblock {\em arXiv:1802.04376}, 2018.

\bibitem{inoue2018data}
Hiroshi Inoue.
\newblock Data augmentation by pairing samples for images classification.
\newblock {\em arXiv:1801.02929}, 2018.

\bibitem{ioffe2015batch}
Sergey Ioffe and Christian Szegedy.
\newblock Batch normalization: Accelerating deep network training by reducing
  internal covariate shift.
\newblock In {\em ICML}, 2015.

\bibitem{joachims1999transductive}
Thorsten Joachims.
\newblock Transductive inference for text classification using support vector
  machines.
\newblock In {\em ICML}, volume~99, pages 200--209, 1999.

\bibitem{kingma2014adam}
Diederik~P Kingma and Jimmy Ba.
\newblock Adam: A method for stochastic optimization.
\newblock In {\em ICLR}, 2015.

\bibitem{krizhevsky2009learning}
Alex Krizhevsky and Geoffrey Hinton.
\newblock Learning multiple layers of features from tiny images.
\newblock Technical report, University of Toronto, 2009.

\bibitem{krizhevsky2012imagenet}
Alex Krizhevsky, Ilya Sutskever, and Geoffrey~E Hinton.
\newblock Imagenet classification with deep convolutional neural networks.
\newblock In {\em NIPS}, pages 1097--1105, 2012.

\bibitem{lee2018gradient}
Yoonho Lee and Seungjin Choi.
\newblock Gradient-based meta-learning with learned layerwise metric and
  subspace.
\newblock In {\em ICML}, 2018.

\bibitem{li2017meta}
Zhenguo Li, Fengwei Zhou, Fei Chen, and Hang Li.
\newblock Meta-sgd: Learning to learn quickly for few shot learning.
\newblock {\em arXiv:1707.09835}, 2017.

\bibitem{Liu2018}
Yanbin Liu, Juho Lee, Minseop Park, Saehoon Kim, Eunho Yang, Sung~Ju Hwang, and
  Yi Yang.
\newblock Learning to propagate labels: Transductive propagation network for
  few-shot learning.
\newblock In {\em ICLR}, 2018.

\bibitem{munkhdalai2017meta}
Tsendsuren Munkhdalai and Hong Yu.
\newblock Meta networks.
\newblock In {\em ICML}, 2017.

\bibitem{munkhdalai2018rapid}
Tsendsuren Munkhdalai, Xingdi Yuan, Soroush Mehri, and Adam Trischler.
\newblock Rapid adaptation with conditionally shifted neurons.
\newblock In {\em ICML}, pages 3661--3670, 2018.

\bibitem{nichol2018first}
Alex Nichol, Joshua Achiam, and John Schulman.
\newblock On first-order meta-learning algorithms.
\newblock {\em CoRR, abs/1803.02999}, 2, 2018.

\bibitem{oreshkin2018tadam}
Boris Oreshkin, Pau~Rodr{\'\i}guez L{\'o}pez, and Alexandre Lacoste.
\newblock Tadam: Task dependent adaptive metric for improved few-shot learning.
\newblock In {\em NIPS}, pages 719--729, 2018.

\bibitem{qiao2017few}
Siyuan Qiao, Chenxi Liu, Wei Shen, and Alan Yuille.
\newblock Few-shot image recognition by predicting parameters from activations.
\newblock In {\em CVPR}, volume~2, 2017.

\bibitem{ravi2016optimization}
Sachin Ravi and Hugo Larochelle.
\newblock Optimization as a model for few-shot learning.
\newblock {\em ICLR}, 2016.

\bibitem{ren2018meta}
Mengye Ren, Eleni Triantafillou, Sachin Ravi, Jake Snell, Kevin Swersky,
  Joshua~B Tenenbaum, Hugo Larochelle, and Richard~S Zemel.
\newblock Meta-learning for semi-supervised few-shot classification.
\newblock In {\em ICLR}, 2018.

\bibitem{rusu2018meta}
Andrei~A Rusu, Dushyant Rao, Jakub Sygnowski, Oriol Vinyals, Razvan Pascanu,
  Simon Osindero, and Raia Hadsell.
\newblock Meta-learning with latent embedding optimization.
\newblock In {\em ICLR}, 2019.

\bibitem{santoro2016meta}
Adam Santoro, Sergey Bartunov, Matthew Botvinick, Daan Wierstra, and Timothy
  Lillicrap.
\newblock Meta-learning with memory-augmented neural networks.
\newblock In {\em ICML}, pages 1842--1850, 2016.

\bibitem{simonyan2014very}
Karen Simonyan and Andrew Zisserman.
\newblock Very deep convolutional networks for large-scale image recognition.
\newblock In {\em ICLR}, 2015.

\bibitem{snell2017prototypical}
Jake Snell, Kevin Swersky, and Richard Zemel.
\newblock Prototypical networks for few-shot learning.
\newblock In {\em NIPS}, pages 4077--4087, 2017.

\bibitem{sturm1999using}
Jos~F Sturm.
\newblock Using sedumi 1.02, a matlab toolbox for optimization over symmetric
  cones.
\newblock {\em Optimization methods and software}, 11(1-4):625--653, 1999.

\bibitem{thrun2012learning}
Sebastian Thrun and Lorien Pratt.
\newblock {\em Learning to learn}.
\newblock Springer Science \& Business Media, 2012.

\bibitem{torresani2007large}
Lorenzo Torresani and Kuang-chih Lee.
\newblock Large margin component analysis.
\newblock In {\em NIPS}, pages 1385--1392, 2007.

\bibitem{triantafillou2017few}
Eleni Triantafillou, Richard Zemel, and Raquel Urtasun.
\newblock Few-shot learning through an information retrieval lens.
\newblock In {\em NIPS}, pages 2255--2265, 2017.

\bibitem{vapnik1999overview}
Vladimir~N Vapnik.
\newblock An overview of statistical learning theory.
\newblock {\em IEEE transactions on neural networks}, 10(5):988--999, 1999.

\bibitem{vinyals2016matching}
Oriol Vinyals, Charles Blundell, Tim Lillicrap, Daan Wierstra, et~al.
\newblock Matching networks for one shot learning.
\newblock In {\em NIPS}, pages 3630--3638, 2016.

\bibitem{wah2011caltech}
Catherine Wah, Steve Branson, Peter Welinder, Pietro Perona, and Serge
  Belongie.
\newblock The caltech-ucsd birds-200-2011 dataset.
\newblock {\em Technical Report CNS-TR-2011-001}, 2011.

\bibitem{weinberger2006distance}
Kilian~Q Weinberger, John Blitzer, and Lawrence~K Saul.
\newblock Distance metric learning for large margin nearest neighbor
  classification.
\newblock In {\em NIPS}, pages 1473--1480, 2006.

\bibitem{yang2018learning}
Flood Sung~Yongxin Yang, Li Zhang, Tao Xiang, Philip~HS Torr, and Timothy~M
  Hospedales.
\newblock Learning to compare: Relation network for few-shot learning.
\newblock In {\em CVPR}, 2018.

\bibitem{zhang2017mixup}
Hongyi Zhang, Moustapha Cisse, Yann~N Dauphin, and David Lopez-Paz.
\newblock mixup: Beyond empirical risk minimization.
\newblock In {\em ICLR}, 2018.

\bibitem{zhou2018deep}
Fengwei Zhou, Bin Wu, and Zhenguo Li.
\newblock Deep meta-learning: Learning to learn in the concept space.
\newblock {\em arXiv:1802.03596}, 2018.

\end{thebibliography}
}

\end{document}